\documentclass{article}
\usepackage{spconf,amsmath,graphicx}
\usepackage{tabularx, booktabs} 
\usepackage{siunitx}            
\usepackage{booktabs}
\usepackage{amssymb}
\usepackage[table]{xcolor}

\usepackage{amsmath}
\usepackage{graphicx}
\title{GRPOformer: Advancing Hyperparameter Optimization via Group Relative Policy Optimization}
\name{Haoxin Guo\textsuperscript{1} Jiawen Pan\textsuperscript{1,2} Weixin Zhai\textsuperscript{1,2*}\thanks{*Corresponding author. }\thanks{ 
This research was financially supported by the National Natural Science Foundation of China (32301691), National Key R\&D Program of China (2025YFE0103600).}}  
\address{\textsuperscript{1}College of Information and Electrical Engineering, \\China Agricultural University, Beijing 100083, China}
\address{\textsuperscript{1}College of Information and Electrical Engineering, \\China Agricultural University, Beijing 100083, China\\\textsuperscript{2}Key Laboratory of Agricultural Machinery Monitoring and Big Data Application, \\Ministry of Agriculture and Rural Affairs, Beijing 100083, China}

\begin{document}
%
\maketitle
\begin{abstract}
Hyperparameter optimization (HPO) plays a critical role in improving model performance. Transformer-based HPO methods have shown great potential; however, existing approaches rely heavily on large-scale historical optimization trajectories and lack effective reinforcement learning (RL) techniques, thereby limiting their efficiency and performance improvements. Inspired by the success of \textbf{Group Relative Policy Optimization (GRPO)} in large language models (LLMs), we propose \textbf{GRPOformer} — a novel hyperparameter optimization framework that integrates reinforcement learning (RL) with \textbf{Transformers}. In GRPOformer, Transformers are employed to generate new hyperparameter configurations from historical optimization trajectories, while GRPO enables rapid trajectory construction and optimization strategy learning from scratch. Moreover, we introduce \textbf{Policy Churn Regularization (PCR)} to enhance the stability of GRPO training. Experimental results on \textbf{OpenML} demonstrate that GRPOformer consistently outperforms baseline methods across diverse tasks, offering new insights into the application of RL for HPO.
\end{abstract}
\section{Introduction}
\label{sec:intro}
Hyperparameter Optimization (HPO) is critical for machine learning (ML) model development\cite{hpoml,yang2020hyperparameter} and widely used in fields like signal processing and control systems\cite{surianarayanan2023survey,firmin2024parallel}. With the advancement of complex ML models (e.g., deep neural networks, large language models), traditional HPO methods (e.g., grid search, Bayesian optimization) face bottlenecks in efficiency and scalability\cite{bischl2023hyperparameter}. To address this, Transformer-based HPO algorithms have emerged, leveraging their sequence modeling capability to capture correlations between historical optimization trajectory and hyperparameter performance~\cite{optformer,optformer-m,llmofhpo}, thereby formulating HPO as a sequential decision task. Building on this paradigm, some studies further integrate reinforcement learning (RL) to enhance optimization adaptability~\cite{24iot,boformer}. This technical paradigm has promoted HPO’s application in scenarios such as deep neural network tuning (computer vision) and large language model fine-tuning (natural language processing), where efficient and adaptive HPO remains a core prerequisite for deploying high-performance models\cite{liu2024language}.

Although Transformer-based hyperparameter optimization (HPO) methods have shown great potential, their reliance on large-scale historical optimization trajectories results in inefficiency, particularly when starting from scratch. Inspired by the successful application of \textbf{Group Relative Policy Optimization (GRPO)} in large language models (LLMs) \cite{grpo}, we introduce GRPO as a reinforcement learning (RL) approach to address this limitation. Building on this, we propose a novel hyperparameter optimization framework that integrates RL with Transformers --- \textbf{GRPOformer}. To the best of our knowledge, this is the first application of GRPO in the HPO setting.

\begin{figure*}[t]  
    \centering
    \includegraphics[width=0.95\textwidth]{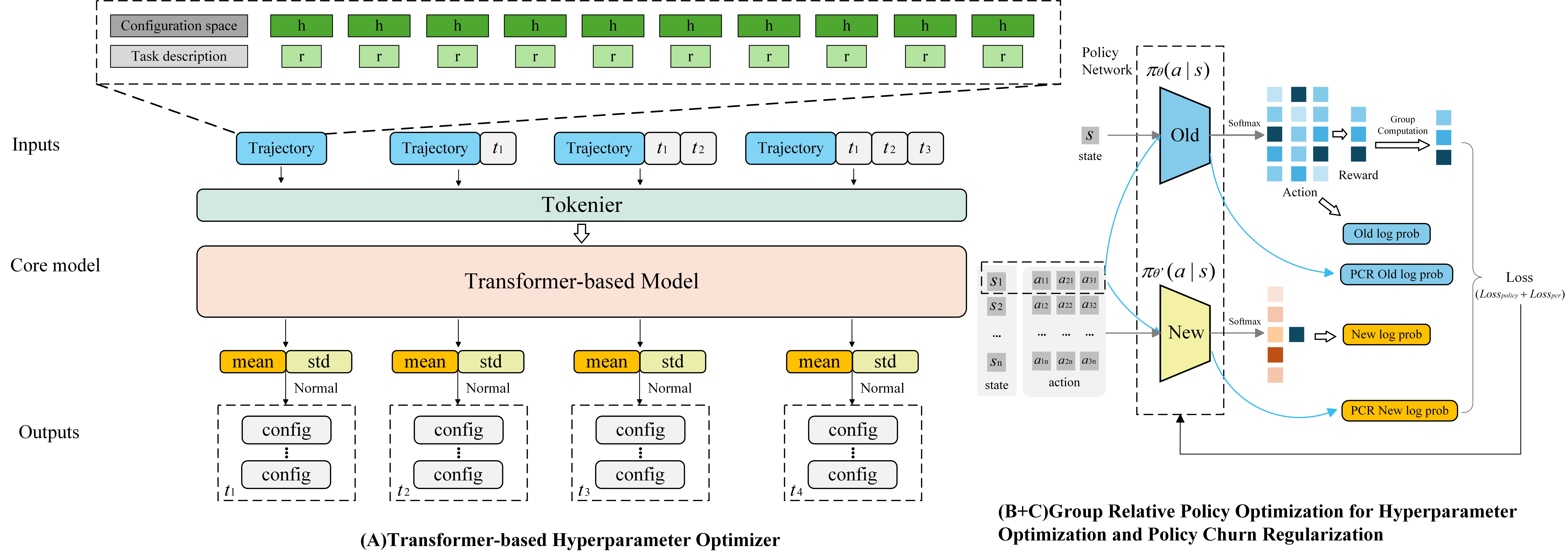}  
    \caption{\textbf{Overall Framework of GRPOformer. }\textbf{(A) Transformer-based Hyperparameter Optimizer:} Input historical optimization trajectory into Transformer to generate new configurations iteratively. \textbf{(B) Group Relative Policy Optimization for HPO:} Critic-free; generate multiple groups per iteration and optimize via inter-group relative advantages. \textbf{(C) Policy Churn Regularization:} Constrain policies with partial historical optimization trajectory to reduce oscillation and improve stability.}
    \label{fig:framework}
\end{figure*}

At its core, GRPOformer integrates the sequence modeling capability of Transformers with the stable optimization mechanism of \textbf{GRPO}. Specifically, we treat the historical optimization trajectory as a learnable sequence signal and feed it into a Transformer, which automatically captures the underlying dependencies between parameter adjustments and performance variations to generate task-adaptive configurations. Meanwhile, by leveraging GRPO's within-group relative-advantage normalization, our framework can compare and update multiple candidate configurations generated in each iteration without relying on an explicitly trained value (critic) network, and integrate the results back into the trajectory for efficient and stable policy optimization. Furthermore, inspired by recent studies \cite{chain}, and to address the issue of abrupt fluctuations during policy updates that can lead to unstable training and hinder convergence, we propose a new regularization strategy — \textbf{Policy Churn Regularization (PCR)}. During each update, PCR incorporates the earlier portion of the trajectory into optimization, constraining policy outputs in those states to prevent sudden shifts, thereby reducing training oscillations and significantly improving convergence stability and efficiency.

To evaluate the effectiveness of the proposed method, we conducted experiments on the \textbf{OpenML} platform\cite{bischl2025openml}. OpenML is an open-source platform that provides standardized datasets and integrates popular machine learning libraries, offering strong support for research and practical applications. Moreover, several benchmark frameworks can readily leverage OpenML datasets for hyperparameter optimization\cite{hpob,hpobench,carps}, facilitating comprehensive evaluation of our experiments and model performance.

 Our contributions are: 1.Motivated by the observation that existing \textbf{Transformer-based} hyperparameter optimization (HPO) algorithms heavily rely on large-scale historical optimization trajectories, we propose \textbf{GRPOformer}, a hyperparameter optimization framework that integrates reinforcement learning with Transformers. This framework formulates HPO as a sequence prediction task and leverages historical optimization trajectories to generate more plausible new configurations.
 2.By adopting \textbf{GRPO}, our framework can generate multiple configurations in parallel and update them using relative advantages, enabling efficient trajectory construction and policy optimization. 3.To mitigate policy churn in GRPO, which undermines efficiency and convergence, we propose a \textbf{PCR} strategy. PCR stabilizes training and accelerates convergence. 4.Validation on real-world \textbf{OpenML} datasets shows that GRPOformer consistently outperforms existing methods.
\section{METHODOLOGY}
\label{sec:meth}
As shown in Fig.~\ref {fig:framework}, the \textbf{GRPOformer} framework comprises three core components: a Transformer-based hyperparameter optimizer for modeling historical optimization trajectories, a GRPO optimization mechanism designed specifically for HPO, and a PCR strategy to stabilize GRPO's optimization process. The following elaborates on these three modules.
\subsection{Transformer-based Hyperparameter Optimizer}
\label{sec:2.1}
In GRPOformer, the core generator for hyperparameter configurations is built upon a \textbf{Transformer-based model}. 
Specifically, the model encodes the task information $\tau$, the hyperparameter configuration space $\mathcal{H}$ 
(including the hyperparameters, their types, and domains), and the historical optimization trajectory $\mathcal{T}$ 
into token sequences that serve as input to the Transformer, enabling joint modeling of task-specific context and historical evaluations. 
Formally, the trajectory is represented as:
\begin{equation}
\mathcal{T} = \{(h_1, r_1), (h_2, r_2), \dots, (h_t, r_t)\},
\end{equation}
where $h_i$ denotes the $i$-th hyperparameter configuration and $r_i$ the corresponding evaluation result.

On top of this representation, the Transformer models the conditional distribution of the next candidate configuration:
\begin{equation}
p(h_{t+1} \mid \mathcal{T}, \tau, \mathcal{H}) = \mathcal{F}_\theta(\mathcal{T}, \tau, \mathcal{H}),
\end{equation}
where $\mathcal{F}_\theta$ denotes a Transformer parameterized by $\theta$. 
In this formulation, hyperparameter optimization in GRPOformer is cast as a conditional sequence prediction problem, 
where new candidates are iteratively generated by leveraging both task information and accumulated historical feedback.
\subsection{Group Relative Policy Optimization for HPO}
\label{sec:2.2}
To address the issues of slow historical optimization trajectory construction and suboptimal optimization accuracy in Transformer-based hyperparameter optimization (HPO) methods, we introduce \textbf{Group Relative Policy Optimization (GRPO)} as the reinforcement learning optimization method in our model. GRPO is a member of the Proximal Policy Optimization (PPO)-like family of algorithms. Unlike standard PPO, which relies on a value function to estimate advantages, GRPO computes relative advantages via intra-group comparisons, eliminating the need for a critic network.

Specifically, given a set of candidate actions $\{a_1, a_2, \dots, a_K\}$ with corresponding rewards 
$r(a_k)$, GRPO defines the relative advantage as:
\begin{equation}
A(a_k) = r(a_k) - \frac{1}{K} \sum_{j=1}^K r(a_j),
\end{equation}
where $A(a_k)$ denotes the relative advantage of action $a_k$, $r(a_k)$ is its reward, $K$ is the 
number of actions in the group, and $\frac{1}{K}\sum_{j=1}^K r(a_j)$ is the group average reward. 
This formula captures how each action deviates from the group mean.

During policy updates, GRPO employs a clipped surrogate objective similar to PPO:
\begin{multline}
\mathcal{L}_{\text{GRPO}}(\theta) = \mathbb{E}\Bigg[ \min \Big( 
\frac{\pi_\theta(a|s)}{\pi_{\theta_{\text{old}}}(a|s)} A(a), \\
\text{clip}\Big(\frac{\pi_\theta(a|s)}{\pi_{\theta_{\text{old}}}(a|s)}, 
1-\epsilon, 1+\epsilon\Big) A(a)
\Big) \Bigg],
\end{multline}
where $\mathcal{L}_{\text{GRPO}}(\theta)$ is the loss function for policy update, 
$\pi_\theta(a|s)$ is the probability of selecting action $a$ in state $s$ under the current policy, 
$\pi_{\theta_{\text{old}}}(a|s)$ is the corresponding probability under the old policy, 
$A(a)$ is the relative advantage, and $\epsilon$ is the clipping parameter. This design ensures 
stable updates while removing the need for a critic network.

In hyperparameter optimization, the application of GRPO can be expressed as: at each iteration, a 
group of candidate configurations $\{h_1, h_2, \dots, h_K\}$ is generated and their evaluation results 
$\{r(h_1), r(h_2), \dots, r(h_K)\}$ are incorporated into the historical optimization trajectory:
\begin{equation}
\mathcal{T}_{t+1} = \mathcal{T}_t \cup \{(h_k, r(h_k))\}_{k=1}^K,
\end{equation}
where $\mathcal{T}_t$ denotes the trajectory at iteration $t$, and $(h_k, r(h_k))$ represents a 
candidate configuration and its evaluation. Once multiple groups are collected, policy optimization 
is performed using inter-group relative advantages, achieving a balance between search efficiency 
and model accuracy.

\subsection{Policy Churn Regularization}
\label{sec:2.3}
In GRPO, policy updates can fluctuate drastically, causing significant changes in action 
probabilities, which may reduce training efficiency and hinder convergence. To address this 
issue and promote smoother policy evolution, we introduce \textbf{Policy Churn Regularization (PCR)} 
in our model to enhance the reinforcement learning performance of GRPO.

Specifically, considering the intrinsic characteristics of the Transformer \cite{wu2025emergence}, we take the earlier portion of the 
historical optimization trajectory as the input state set $s$, and measure the divergence between the old policy 
$\pi_{\text{old}}$ and the updated policy $\pi_\theta$. The regularization loss is defined as the KL divergence:
\begin{equation}
    L_{\text{PC}} = \mathbb{E}_{s \sim D} \left[ D_{\text{KL}}\left( \pi_{\text{old}}(\cdot \mid s) \, \| \, \pi_{\theta}(\cdot \mid s) \right) \right],
\end{equation}
where $D$ denotes the set of reference states, penalizing large deviations between consecutive policies.

The final GRPO policy loss is:
\begin{equation}
    L_{\text{policy}} = L_{\text{GRPO}} + \lambda_{\text{PC}} \, L_{\text{PC}},
\end{equation}
ensuring that GRPO maintains the benefits of group-wise relative optimization while allowing smoother 
policy evolution.

\section{EXPERIMENTS}
\label{sec:exp}
\subsection{Experimental Setup}
\label{sec:3.1}
\textbf{Datasets and Baselines. }We conducted experiments on the OpenML platform using 6 representative machine learning models: Extreme Gradient Boosting (XGBoost), Support Vector Machine (SVM), Logistic Regression (LR), Random Forest (RF), Neural Network (NN), and Histogram-based Gradient Boosting (HistGB). Across 6 datasets, we constructed 36 hyperparameter optimization scenarios. We benchmark against recent SOTA methods, including BOform\-er\cite{boformer},LLM-HPO\cite{llmofhpo}, OPTformer\cite{optformer}, SBOA\cite{SOBA}.\\
\textbf{Evaluation Metrics. }To evaluate algorithm superiority, accuracy is used as the core metric, along with a normalized metric. The normalized metric is calculated as the best normalized accuracy at the $t$-th trial: $\max_{i \in \{1:t\}} \frac{y_i - y_{\text{rand}}}{y_{\text{max}} - y_{\text{rand}}}$, where $y_i$ is the accuracy at the $i$-th trial, $y_{\text{rand}}$ is the benchmark derived from 500 random configurations per task, and $y_{\text{max}}$ is the best accuracy found by all algorithms. Evaluation metrics across all experiments include: Beat the Random (BtR, proportion of tasks where the target method outperforms random search, higher is better), Median Performance (MP, median of normalized performance metrics of the target method relative to random search, higher is better), Mean Performance (MnP, mean of normalized performance metrics of the target method relative to random search, higher is better), and Mean Rank (MnR, comprehensive ranking of the target method against other competing methods, lower is better).
\begin{figure*}[t]  
    \centering
    \includegraphics[width=0.83\textwidth]{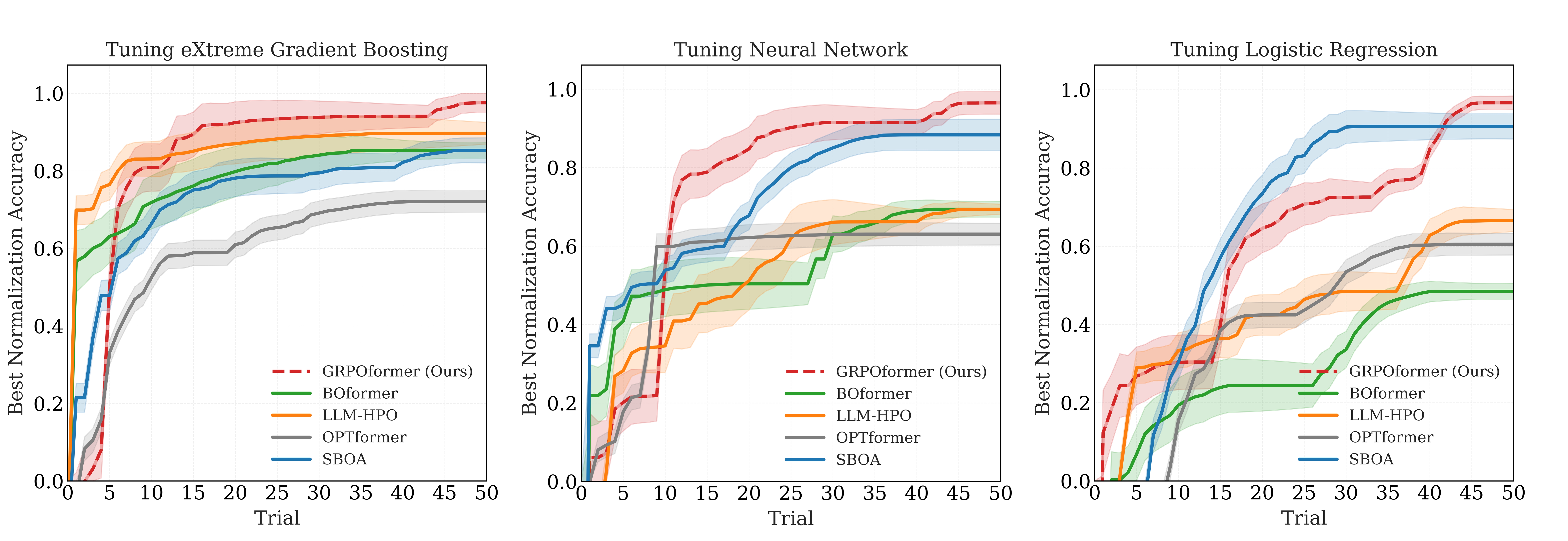}  
    \caption{Comparison Chart of Comparative Experiment Results Across Three Scenarios}
    \label{fig:experiments}
\end{figure*}
\subsection{Main Results and Analysis}
\label{sec:3.2}

Table \ref{tab:Comparative Experiment} presents the results of GRPOformer against several mainstream HPO methods, including Transformer-based approaches (OPTformer, LLM-HPO), reinforcement learning with Transformer (BOformer), and a widely used metaheuristic algorithm (SBOA). The results show that GRPOformer consistently outperforms all baselines across all metrics: it achieves 94.44\% on BtR, surpassing BOformer (88.89\%), LLM-HPO (83.33\%), OPTformer (77.78\%), and SBOA (91.67\%); obtains the highest scores on MP (0.9545) and MnP (0.9187); and ranks first with the lowest MnR value of 1.81. These findings indicate that GRPOformer delivers superior efficiency, performance stability, and overall ranking compared to existing methods.

\begin{table}[ht]
\centering
\caption{Comparative Experiments of GRPOformer}
\label{tab:Comparative Experiment}
\small
\begin{tabularx}{\columnwidth}{lXXXX} 
\toprule
\textbf{Model} & \textbf{BtR$\uparrow$} & \textbf{MP$\uparrow$} & \textbf{MnP$\uparrow$} & \textbf{MnR$\downarrow$} \\
\midrule
\rowcolor{gray!10}
\textbf{GRPOformer} & \textbf{94.44\%} & \textbf{0.9545} & \textbf{0.9187} & \textbf{1.81} \\
BOformer   & 88.89\% & 0.6437 & 0.5767 & 3.14 \\
LLM-HPO    & 83.33\% & 0.3224 & 0.0863 & 3.83 \\
OPTformer  & 77.78\% & 0.5882 & 0.4360 & 3.75 \\
SBOA       & 91.67\% & 0.8983 & 0.8678 & 1.92 \\
\bottomrule
\end{tabularx}
\end{table}
To provide a more intuitive demonstration of the superiority of our algorithm, we plot the optimization curves in Fig.~\ref{fig:experiments}, where the horizontal axis denotes the number of trials and the vertical axis represents the best normalization accuracy (restricted to values greater than zero). As shown, across all three scenarios, GRPOformer consistently achieves the best performance, with its curve rising more sharply and reaching higher accuracy levels compared to the baselines.

\subsection{Ablation Study}
\label{sec:3.3}

As shown in Table~\ref{tab:Melting Experiment}, we designed two ablation settings: (1) removing PCR while retaining the original GRPO update rule (\textbf{w/o PCR}); and (2) removing GRPO, where all generated configurations are directly traced back to the trajectory and used for the next update (\textbf{w/o GRPO}). The results demonstrate the effectiveness of both the GRPO mechanism and the Policy Churn Regularization (PCR) in GRPOformer. Removing PCR (w/o PCR) leads to a moderate drop in performance, with BtR decreasing from 94.44\% to 91.67\%, MP from 0.9545 to 0.8332, MnP from 0.9187 to 0.7982, and MnR increasing from 1.31 to 1.97. Removing GRPO (w/o GRPO) significantly degrades the performance, resulting in BtR of 83.33\%, MP of 0.5143, MnP of 0.5260, and MnR of 2.78. These results indicate that both GRPO and PCR are crucial components contributing to the superior performance of GRPOformer.

\begin{table}[ht]
\centering
\caption{Ablation Study Results of GRPOformer}
\label{tab:Melting Experiment}
\small
\begin{tabularx}{\columnwidth}{lXXXX} 
\toprule
\textbf{Model} & \textbf{BtR$\uparrow$} & \textbf{MP$\uparrow$} & \textbf{MnP$\uparrow$} & \textbf{MnR$\downarrow$} \\
\midrule
\rowcolor{gray!10}
\textbf{GRPOformer} & \textbf{94.44\%} & \textbf{0.9545} & \textbf{0.9187} & \textbf{1.31} \\
w/o PCR  & 91.67\% & 0.8332  & 0.7982 & 1.97 \\
w/o GRPO  & 83.33\% & 0.5143   & 0.5260   & 2.78 \\
\bottomrule
\end{tabularx}
\end{table}

\section{CONCLUSION}
\label{sec:con}

In this work, we propose \textbf{GRPOformer}, a hyperparameter optimization framework built upon 
\textbf{Group Relative Policy Optimization (GRPO)}, designed to address the limitations of existing 
\textbf{Transformer-based} hyperparameter optimizers. At its core, the framework employs a Transformer 
architecture that estimates hyperparameter configurations by processing accumulated historical 
optimization trajectory. In each iteration, the model generates multiple candidate configurations, 
integrates them by computing relative advantages, and drives the GRPO learning process, thereby 
optimizing the core model via reinforcement learning. Furthermore, we incorporate 
\textbf{Policy Churn Regularization (PCR)} to alleviate drastic fluctuations during policy updates, 
ensuring more stable optimization and enhancing RL performance. Extensive comparative experiments 
and ablation studies demonstrate the effectiveness of our approach and highlight a promising 
direction for applying RL in hyperparameter optimization.

\bibliographystyle{IEEEbib}
\bibliography{refs}

\begin{thebibliography}{10}
\providecommand{\url}[1]{#1}
\csname url@samestyle\endcsname
\providecommand{\newblock}{\relax}
\providecommand{\bibinfo}[2]{#2}
\providecommand{\BIBentrySTDinterwordspacing}{\spaceskip=0pt\relax}
\providecommand{\BIBentryALTinterwordstretchfactor}{4}
\providecommand{\BIBentryALTinterwordspacing}{\spaceskip=\fontdimen2\font plus
\BIBentryALTinterwordstretchfactor\fontdimen3\font minus \fontdimen4\font\relax}
\providecommand{\BIBforeignlanguage}[2]{{%
\expandafter\ifx\csname l@#1\endcsname\relax
\typeout{** WARNING: IEEEtran.bst: No hyphenation pattern has been}%
\typeout{** loaded for the language `#1'. Using the pattern for}%
\typeout{** the default language instead.}%
\else
\language=\csname l@#1\endcsname
\fi
#2}}
\providecommand{\BIBdecl}{\relax}
\BIBdecl

\bibitem{chain}
\BIBentryALTinterwordspacing
H.~Tang and G.~Berseth, ``Improving deep reinforcement learning by reducing the chain effect of value and policy churn,'' in \emph{Advances in Neural Information Processing Systems}, 2024. [Online]. Available: \url{https://openreview.net/pdf?id=cQoAgPBARc}
\BIBentrySTDinterwordspacing

\bibitem{boformer}
Y.-H. Hung, K.-J. Lin, Y.-H. Lin, C.-Y. Wang, C.~Sun, and P.-C. Hsieh, ``Boformer: Learning to solve multi-objective bayesian optimization via non-markovian rl,'' \emph{arXiv preprint arXiv:2505.21974}, 2025.

\bibitem{optformer}
Y.~Chen, X.~Song, C.~Lee, Z.~Wang, R.~Zhang, D.~Dohan, K.~Kawakami, G.~Kochanski, A.~Doucet, M.~Ranzato \emph{et~al.}, ``Towards learning universal hyperparameter optimizers with transformers,'' \emph{Advances in Neural Information Processing Systems}, vol.~35, pp. 32\,053--32\,068, 2022.

\bibitem{24iot}
I.~Shaer, S.~Nikan, and A.~Shami, ``Efficient transformer-based hyper-parameter optimization for resource-constrained iot environments,'' \emph{IEEE Internet of Things Magazine}, vol.~7, no.~6, pp. 102--108, 2024.

\bibitem{SOBA}
Y.~Fu, D.~Liu, J.~Chen, and L.~He, ``Secretary bird optimization algorithm: a new metaheuristic for solving global optimization problems,'' \emph{Artificial Intelligence Review}, vol.~57, no.~5, p. 123, 2024.

\bibitem{hpob}
S.~P. Arango, H.~S. Jomaa, M.~Wistuba, and J.~Grabocka, ``Hpo-b: A large-scale reproducible benchmark for black-box hpo based on openml,'' \emph{arXiv preprint arXiv:2106.06257}, 2021.

\bibitem{hpobench}
K.~Eggensperger, P.~M{\"u}ller, N.~Mallik, M.~Feurer, R.~Sass, A.~Klein, N.~Awad, M.~Lindauer, and F.~Hutter, ``Hpobench: A collection of reproducible multi-fidelity benchmark problems for hpo,'' \emph{arXiv preprint arXiv:2109.06716}, 2021.

\bibitem{carps}
C.~Benjamins, H.~Graf, S.~Segel, D.~Deng, T.~Ruhkopf, L.~Hennig, S.~Basu, N.~Mallik, E.~Bergman, D.~Chen \emph{et~al.}, ``carps: A framework for comparing n hyperparameter optimizers on m benchmarks,'' \emph{arXiv preprint arXiv:2506.06143}, 2025.

\bibitem{grpo}
Z.~Shao, P.~Wang, Q.~Zhu, R.~Xu, J.~Song, X.~Bi, H.~Zhang, M.~Zhang, Y.~Li, Y.~Wu \emph{et~al.}, ``Deepseekmath: Pushing the limits of mathematical reasoning in open language models,'' \emph{arXiv preprint arXiv:2402.03300}, 2024.

\bibitem{surianarayanan2023survey}
C.~Surianarayanan, J.~J. Lawrence, P.~R. Chelliah, E.~Prakash, and C.~Hewage, ``A survey on optimization techniques for edge artificial intelligence (ai),'' \emph{Sensors}, vol.~23, no.~3, p. 1279, 2023.

\bibitem{optformer-m}
L.~M. Dery, A.~L. Friesen, N.~De~Freitas, M.~Ranzato, and Y.~Chen, ``Multi-step planning for automated hyperparameter optimization with optformer,'' \emph{arXiv preprint arXiv:2210.04971}, 2022.

\bibitem{hpoml}
L.~Franceschi, M.~Donini, V.~Perrone, A.~Klein, C.~Archambeau, M.~Seeger, M.~Pontil, and P.~Frasconi, ``Hyperparameter optimization in machine learning,'' \emph{arXiv preprint arXiv:2410.22854}, 2024.

\bibitem{bischl2023hyperparameter}
B.~Bischl, M.~Binder, M.~Lang, T.~Pielok, J.~Richter, S.~Coors, J.~Thomas, T.~Ullmann, M.~Becker, A.-L. Boulesteix \emph{et~al.}, ``Hyperparameter optimization: Foundations, algorithms, best practices, and open challenges,'' \emph{Wiley Interdisciplinary Reviews: Data Mining and Knowledge Discovery}, vol.~13, no.~2, p. e1484, 2023.

\bibitem{llmofhpo}
M.~R. Zhang, N.~Desai, J.~Bae, J.~Lorraine, and J.~Ba, ``Using large language models for hyperparameter optimization,'' \emph{arXiv preprint arXiv:2312.04528}, 2023.

\bibitem{liu2024language}
S.~Liu, S.~Yu, Z.~Lin, D.~Pathak, and D.~Ramanan, ``Language models as black-box optimizers for vision-language models,'' in \emph{Proceedings of the IEEE/CVF Conference on Computer Vision and Pattern Recognition}, 2024, pp. 12\,687--12\,697.

\bibitem{wu2025emergence}
X.~Wu, Y.~Wang, S.~Jegelka, and A.~Jadbabaie, ``On the emergence of position bias in transformers,'' \emph{arXiv preprint arXiv:2502.01951}, 2025.

\bibitem{firmin2024parallel}
T.~Firmin, P.~Boulet, and E.-G. Talbi, ``Parallel hyperparameter optimization of spiking neural networks,'' \emph{Neurocomputing}, vol. 609, p. 128483, 2024.

\bibitem{bischl2025openml}
B.~Bischl, G.~Casalicchio, T.~Das, M.~Feurer, S.~Fischer, P.~Gijsbers, S.~Mukherjee, A.~C. M{\"u}ller, L.~N{\'e}meth, L.~Oala \emph{et~al.}, ``Openml: Insights from 10 years and more than a thousand papers,'' \emph{Patterns}, 2025.

\bibitem{yang2020hyperparameter}
L.~Yang and A.~Shami, ``On hyperparameter optimization of machine learning algorithms: Theory and practice,'' \emph{Neurocomputing}, vol. 415, pp. 295--316, 2020.

\end{thebibliography}
\end{document}